\documentclass[letterpaper]{article} 
\usepackage{aaai24}  
\usepackage{times}  
\usepackage{helvet}  
\usepackage{courier}  
\usepackage[hyphens]{url}  
\usepackage{graphicx} 
\usepackage{natbib}  
\usepackage{caption} 
\usepackage{algorithm}
\usepackage[noend]{algpseudocode}
\usepackage{multirow}
\usepackage{tabularx}
\usepackage[flushleft]{threeparttable}
\usepackage{makecell}
\usepackage{booktabs}
\usepackage{amsmath}
\usepackage{amssymb}

\usepackage{newfloat}
\usepackage{listings}
\DeclareCaptionStyle{ruled}{labelfont=normalfont,labelsep=colon,strut=off} 
\floatstyle{ruled}
\newfloat{listing}{tb}{lst}{}
\floatname{listing}{Listing}
\title{Transformer-based Selective Super-Resolution for Efficient Image Refinement}
\author{
    Tianyi Zhang\textsuperscript{\rm 1}, 
    Kishore Kasichainula\textsuperscript{\rm 2},
    Yaoxin Zhuo\textsuperscript{\rm 2},
    Baoxin Li\textsuperscript{\rm 2},
    Jae-Sun Seo\textsuperscript{\rm 3},
    Yu Cao\textsuperscript{\rm 1}
}
\affiliations{
    \textsuperscript{\rm 1}University of Minnesota\\
    \textsuperscript{\rm 2}Arizona State University\\
    \textsuperscript{\rm 3}Cornell Tech\\

    zhan9167@umn.edu, \{kkasicha, yzhuo6, baoxin.li\}@asu.edu, js3528@cornell.edu, yucao@umn.edu
}

\usepackage{bibentry}

\begin{document}

\maketitle

\begin{abstract}
Conventional super-resolution methods suffer from two drawbacks: substantial computational cost in upscaling an entire large image, and the introduction of extraneous or potentially detrimental information for downstream computer vision tasks during the refinement of the background. To solve these issues, we propose a novel transformer-based algorithm, Selective Super-Resolution (SSR), which partitions images into non-overlapping tiles, selects tiles of interest at various scales with a pyramid architecture, and exclusively reconstructs these selected tiles with deep features. Experimental results on three datasets demonstrate the efficiency and robust performance of our approach for super-resolution. Compared to the state-of-the-art methods, the FID score is reduced from 26.78 to 10.41 with 40\% reduction in computation cost for the BDD100K dataset.  The source code is available at {https://github.com/destiny301/SSR}.
\end{abstract}

\section{Introduction} \label{sec:intro}
Super-resolution (SR) is a fundamental task aimed at enhancing image resolution by producing intricate details from low-resolution (LR) images. It supplies high-resolution (HR) images that are pivotal for downstream computer vision tasks, such as object detection and image classification, with wide-ranging applications in the real world. For instance, in the context of autonomous driving, higher-resolution images facilitate more precise and early object detection, particularly for diminutive objects. Although various super-resolution methods based on convolutional neural networks (CNNs) have been proposed, which enhance high-frequency information through low-resolution image reconstruction, their efficacy is impeded by a lack of long-range dependency integration.

\begin{figure}[t]
\centering
\includegraphics[width=0.9\columnwidth]{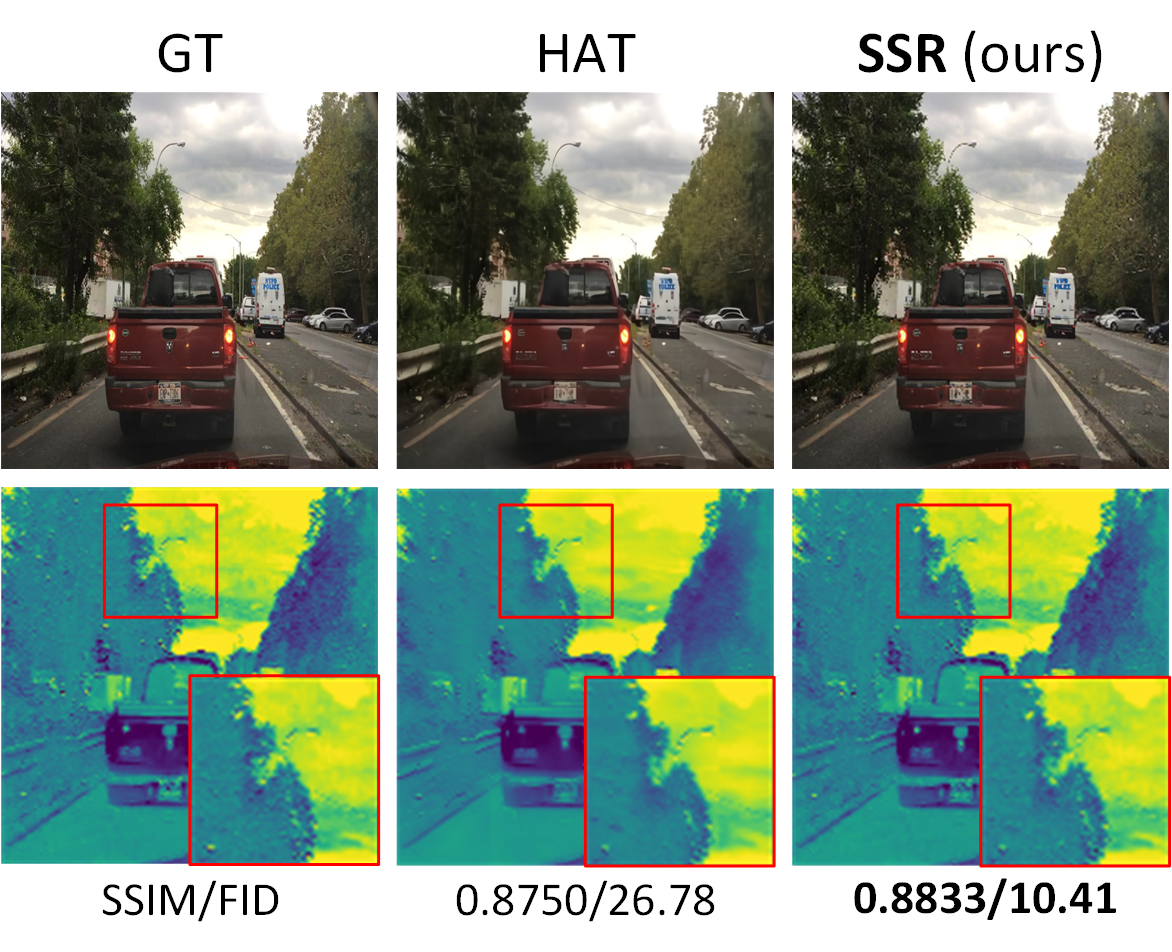} 
\caption{The reconstruction of high-frequency background information in conventional SR methods (e.g., HAT) often results in discrepancies in features compared to ground truth high-resolution (HR) images. SSR effectively resolves this issue by exclusively enhancing foreground pixels.}
\label{fig:intro}
\end{figure}

Recently, leveraging transformer-based architectures to capture the extended contextual information, pioneering efforts like SwinIR \cite{liang2021swinir} and HAT \cite{chen2023activating}, have achieved notable advancements in super-resolution. Nevertheless, two key issues persist with these algorithms. Firstly, due to the substantial scale of transformer-based networks, the computational demand becomes exceedingly high when reconstructing entire images, particularly when input low-resolution image sizes are not small, such as 256$\times$256. Secondly, even in state-of-the-art super-resolution approaches, the refined images fail to match the performance of ground truth HR images for typical downstream tasks.
To delve into the cause of this degradation, we conduct a comparison between features generated by the Inception model from refined images and original HR images. This analysis unveils that features derived from background pixels exhibit markedly different details compared to the ground truth feature map, as depicted in Figure \ref{fig:intro}. This divergence suggests that overemphasizing background pixel details can introduce erroneous information and impede feature generation for downstream tasks.

This paper introduces a novel algorithm, Selective Super-Resolution (SSR), designed to address these challenges. Specifically, leveraging object location information, we partition images into non-overlapping tiles and employ a cost-efficient transformer-based network for tile selection. To ensure comprehensive coverage of objects, a pyramid structure is devised for tile selection across multiple scales. In the final layer of this selection network, we integrate a Gumbel-Softmax layer \cite{jang2016categorical} to make hard decisions, subsequently directing positive tiles to an ensuing super-resolution (SR) module. This module comprises a convolution layer for shallow feature extraction, transformer blocks for deep feature extraction and an image reconstruction block. In contrast, negative tiles are reconstructed directly from shallow features. This framework not only reduces computation by extracting deep features solely for positive tiles but also enhances image generation by avoiding excessive background detail addition.
To validate the robustness of SSR, alongside common evaluation metrics, such as Structural Similarity Index Measure (SSIM), Fréchet Inception Distance (FID), and Kernel Inception Distance (KID), we introduce a novel metric, inspired by KID and tailored to evaluate features from an object detection (OD) model, YOLO, called Kernel YOLO Distance (KYD).
Our approach is experimented on three distinct image datasets, BDD100K, MSRA10K, and COCO2017, demonstrating both substantial computational reduction and image generation improvement.

To summarize, our key contributions are as follows:
\begin{itemize}
    \item We design a low-cost Tile Selection module, employing transformer blocks, to effectively extract desired object-containing tiles from images. The introduction of a pyramid structure ensures accurate selection of positive tiles.
    \item By seamlessly integrating two transformer-based modules, Tile Selection (TS) and Tile Refinement (TR), our SSR efficiently reconstructs the input images by exclusively adding high-frequency information for object-containing tiles, effectively mitigating computational costs and enhancing visual quality.
    \item Through comprehensive experiments on three diverse image datasets and various evaluation metrics, we showcase SSR's robust performance and specifically lower FID from 26.78 to 10.41  for BDD100K, accompanied by a 40\% reduction in computation cost.
\end{itemize}

\section{Related Work}
\subsection{Super-resolution}
In this context, our primary focus revolves around single image super-resolution (SISR) techniques, without delving into methodologies that reconstruct high-resolution images using multiple input images.
Previously, the SRCNN model \cite{dong2014learning} based on convolutional neural networks (CNNs) inspires many works in SR \cite{tai2017image, niu2020single, mei2021image}.
 This seminal approach employed bicubic interpolation and trained a three-layer CNN for SR, achieving remarkable success.
A diverse range of CNN-based methodologies has emerged to map low-resolution images to their high-resolution counterparts, leveraging distinct block designs, such as the incorporation of residual blocks \cite{lim2017enhanced}. Moreover, harnessing the unprecedented image generation capabilities of generative adversarial networks (GANs) \cite{goodfellow2020generative}. Certain studies have notably advanced the quality of generated images, such as SRGAN \cite{ledig2017photo, wang2021real, wang2018esrgan, zhang2019ranksrgan}, which introduce adversarial learning to improve perceptual quality.
Recent strides have been witnessed in the adoption of the attention mechanism, which excels in capturing long-range dependencies. This mechanism has lately been adopted to further enhance SISR methodologies\cite{liang2021swinir, chen2023activating}.

\subsection{Transformer in Computer Vision}
Given the remarkable success of transformers in natural language processing (NLP) \cite{vaswani2017attention}, this architectural paradigm is progressively permeating diverse computer vision tasks \cite{chu2021twins, huang2021shuffle, dong2022cswin, he2022masked, zhang2023improving, zhang2023patch}.  For instance, Vision Transformer (ViT) divides input images into $16\times16$ patches, which are subsequently treated as tokens for the application of the attention mechanism \cite{dosovitskiy2020image}. For object detection (OD), DETR conceptualizes it as a direct set prediction issue and crafts a transformer-based network \cite{carion2020end}. DINO introduces self-supervised learning to propose a novel network rooted in the ViT architecture \cite{caron2021emerging}. Swin Transformer integrates window-based self-attention and shifted window-based self-attention mechanisms, to reduce the computation cost by limiting the computation inside small windows \cite{liu2021swin, liu2022swin}. By capturing long-range dependencies and facilitating enhanced visual representation, transformer-based networks have exhibited superiority in various domains, including super-resolution \cite{liang2021swinir, chen2023activating}.

\section{Methodology}

\begin{figure*}[t]
\centering
\includegraphics[width=0.99\textwidth]{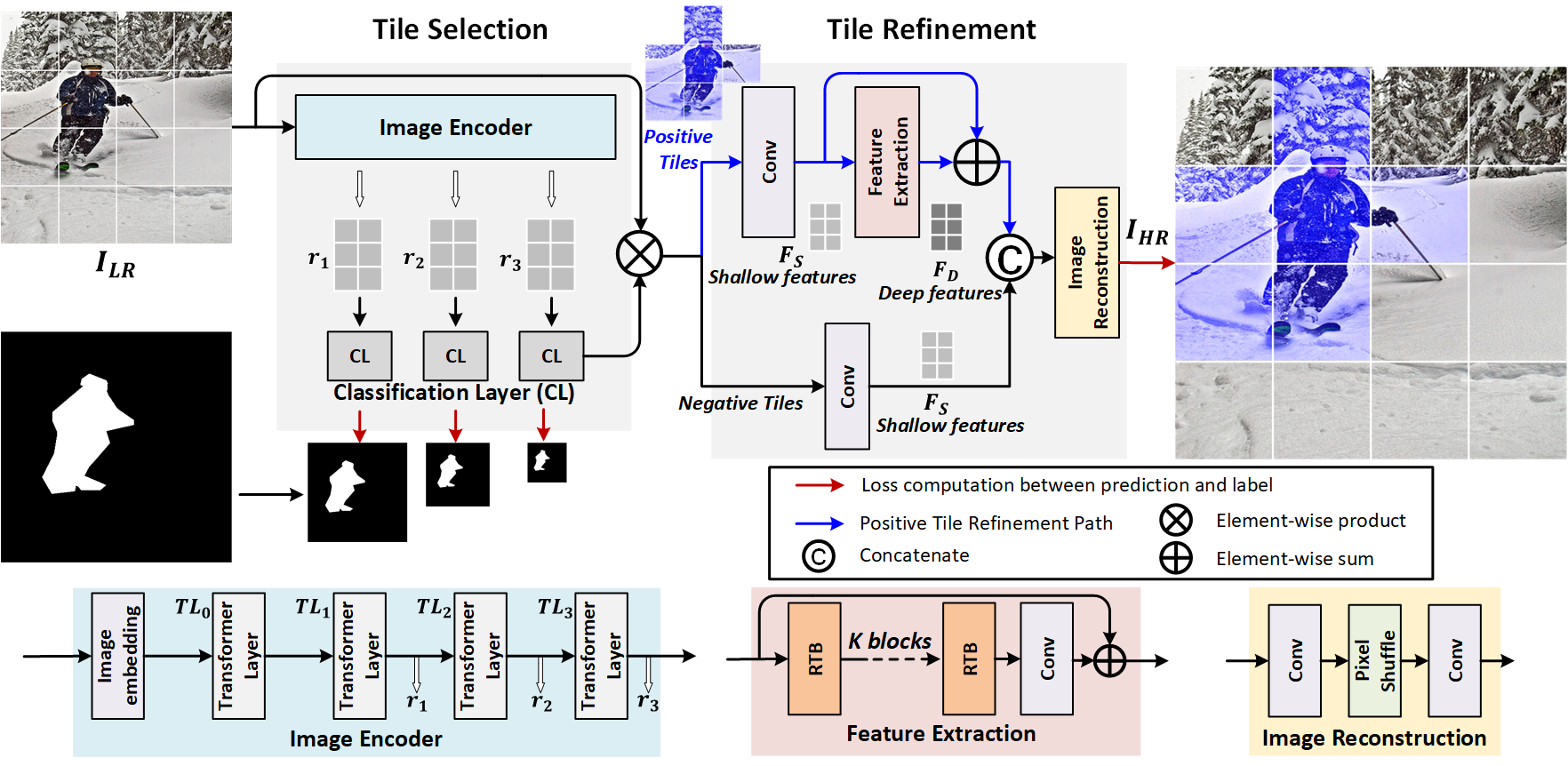} 
\caption{\textbf{Overall architecture of Selective Super-Resolution (SSR).}Input low-resolution (LR) images are split into non-overlapping tiles and classified to two classes by Tile Selection (TS) module. And only positive tiles containing objects are intricately reconstructed by Positive Tile Refinement (PTR) path with deep features. The specific Residual Transformer Block (RTB) in the feature extraction module will be introduced in detail later. }
\label{fig:flowchart}
\end{figure*}

The overall architecture of our approach is illustrated in Figure \ref{fig:flowchart}. Selective Super-Resolution (SSR) network comprises two fundamental modules: Tile Selection (TS) and Tile Refinement (TR).  Both modules are transformer-based networks, with TS being notably smaller in scale.  Upon receiving an input image, the TS module initiates the process by partitioning it into non-overlapping tiles and then selects pertinent tiles based on object location cues. The tiles containing objects are directed through a computationally intensive block for intricate refinement, while the remaining tiles traverse a cost-efficient block for straightforward upscaling. This section is dedicated to a comprehensive discourse on all modules, elucidating their specifics. We outline the precise algorithm in Algorithm \ref{alg:algorithm}.

\subsection{Tile Selection Module}
By splitting the input images into $4\times4$ non-overlapping tiles and utilizing the location information of objects in these images, this module classifies each tile by whether it contains objects or not. Specifically, as shown in Figure \ref{fig:flowchart}, this module consists of an image encoder and several tile classifiers.
First, the embedding layer and the first transformer layer of the encoder embeds the low-resolution (LR) input image $\boldsymbol{I}_{LR}\in\mathbb{R}^{H_{LR}\times W_{LR}\times C_{LR}}$ ($H_{LR}$, $W_{LR}$, and $C_{LR}$ are the input LR image height, width and the number of channels) into representations $r_0\in\mathbb{R}^{\frac{H_{LR}}{p}\times \frac{W_{LR}}{p}\times C}$, where $p$ is the patch size of each token and $C$ is the number of channels for each token.
After that, three transformer layers, $TL_1$, $TL_2$, and $TL_3$, generate representations of these tokens at three different scales.
The transformer layer adopts the structure of Swin Transformer \cite{liu2021swin, liu2022swin}. Basically, with the incorporation of window-based self-attention and shifted window-based self-attention mechanisms, this layer can capture global contextual information with attention computation within small windows, which is much cheaper than the traditional attention mechanism. Besides, with the feature merging layer, we can obtain features at various scales to enable the implementation of pyramid structure. The whole process is as follows,
\begin{equation}
     \boldsymbol{r}_1 = TL_1(\boldsymbol{r}_{0}), \;\;\;
     \boldsymbol{r}_2 = TL_2(\boldsymbol{r}_1), \;\;\;
     \boldsymbol{r}_3 = TL_3(\boldsymbol{r}_2)
\end{equation}

where
$\boldsymbol{r}_1\in\mathbb{R}^{\frac{H_{LR}}{2p}\times \frac{W_{LR}}{2p}\times 2C}$, $\boldsymbol{r}_2\in\mathbb{R}^{\frac{H_{LR}}{4p}\times \frac{W_{LR}}{4p}\times 4C}$, $\boldsymbol{r}_3\in\mathbb{R}^{\frac{H_{LR}}{8p}\times \frac{W_{LR}}{8p}\times 8C}$ are three generated representations for classification. And, each token of these representations corresponds to tiles of size $2p\times 2p$, $4p \times 4p$, and $8p\times 8p$, respectively.

To classify these tiles, we adopt the cross-attention mechanism by introducing a learnable classification token, denoted as $c$. We obtain the query matrix $Q$ by applying one linear layer to image features $r_1$, $r_2$ or $r_3$ as $Q_i=r_iW_i^q, i\in 1, 2, 3$ while computing the key matrix and value matrix with the classification token as equations $K_i=cW_i^k, V_i=cW_i^v, i\in 1, 2, 3$, then the attention computation is expressed as follows,
\begin{equation}
    \boldsymbol{A}_i = softmax(\frac{\boldsymbol{Q}_i\boldsymbol{K}_i^T}{\sqrt{d}})\boldsymbol{V}_i \;\;\; \forall i\in {1, 2, 3}
\end{equation}

Next, each of the three features undergoes individual processing through a multi-layer perceptron (MLP) and a Gumbel-Softmax layer. This step facilitates making definitive classifications for the tile classes. This is crucial for the subsequent refinement module to apply the corresponding network. The process are as follows:
\begin{equation}
    \boldsymbol{s}_i = GumbelSoftmax(MLP_i(\boldsymbol{A}_i)) \;\;\; \forall i\in {1, 2, 3}
\end{equation}
where $MLP_i$ denotes the output layer for the $i$th feature embeddings.

Accordingly to the network structure, after pooling the instance segmentation label of the input images at three different scales, we introduce a pyramid label that contains, $\boldsymbol{y}_1\in\mathbb{R}^{\frac{H_{LR}}{2p}\times \frac{W_{LR}}{2p}\times 1}$, $\boldsymbol{y}_2\in\mathbb{R}^{\frac{H_{LR}}{4p}\times \frac{W_{LR}}{4p}\times 1}$, and $\boldsymbol{y}_3\in\mathbb{R}^{\frac{H_{LR}}{8p}\times \frac{W_{LR}}{8p}\times 1}$, to supervise the training of TS module by allocating positive labels to the tiles that contain objects. This hierarchical structure ensures the preservation of a larger number of tiles and minimizes the loss of positive instances. Finally, all tiles are divided into two groups to be processed by the subsequent refinement module.

\subsection{Tile Refinement Module}
As depicted in Figure \ref{fig:flowchart}, there are two Tile Refinement (TR) paths: Positive Tile Refinement (PTR) targeting object-containing tiles, and Negative Tile Refinement (NTR) for tiles with solely background pixels.

\begin{algorithm}[tb]
\caption{Selective Super-Resolution (SSR)}
\label{alg:algorithm}
\textbf{Input}: Low-resolution (LR) Image data $\boldsymbol{I}_{LR}$, tile class labels $\boldsymbol{y}_1$, $\boldsymbol{y}_2$, $\boldsymbol{y}_3$, and ground truth high-resolution (HR) image $\boldsymbol{I}_{GT}$\\
\textbf{Parameter}: loss weight $\alpha$, learning rate $\eta$\\
\begin{algorithmic}[1] 
\State Initialize model parameters $\theta$
\For{each epoch t = 1, 2, ...}
\State $\boldsymbol{r}_{1}, \boldsymbol{r}_2, \boldsymbol{r}_3 = f_{\theta_{IE}}(\boldsymbol{I}_{LR})$
\Comment{Generate representations with Image Encoder (IE)}
\State $\boldsymbol{s}_i = GumbelSoftmax(f_{\theta_{CL}}(\boldsymbol{r}_i))\;\;\; \forall i\in {1, 2, 3}$ \Comment{Classify tiles with Classification Layer (CL)}
\State $\mathcal{L}_{TS}(\theta) = \sum_{i=1}^3(-\boldsymbol{y}_ilog(\boldsymbol{s}_i) - (1-\boldsymbol{y}_i)log(1-\boldsymbol{s}_i))$ 

\For{each tile $T_{LR}^1, T_{LR}^2, ..., T_{LR}^N$}
\If {$s_3^n = 1$} \Comment{Positive tiles}
\State $F_S^n = f_{\theta_{Conv}}(T_{LR}^n)$ \Comment{Shallow feature extraction with convolution layer}
\State $F_D^n = f_{\theta_{FE}}(F_S^n))$ \Comment{Deep Feature Extraction (FE)}
\State $T_{HR}^n = f_{\theta_{IR}}(F_S^n+F_D^n)$ \Comment{HR Image Reconstruction (IR)}
\Else \Comment{Negative tiles}
\State $F_S^n = f_{\theta_{Conv}}(T_{LR}^n)$
\State $T_{HR}^n = f_{\theta_{IR}}(F_S^n)$
\EndIf
\State \textbf{end if}
\EndFor
\State \textbf{end for}
\State Group all output tiles $T_{HR}$ to entire images $I_{HR}$
\State $\mathcal{L}_{TR}(\theta) = ||I_{HR}-I_{GT}||_1$
\State $\mathcal{L}_{SSR}(\theta) = \mathcal{L}_{TS}(\theta)+\alpha \mathcal{L}_{TR}(\theta)$ 
\State $\theta \leftarrow \theta - \eta \nabla_\theta \mathcal{L}_{SSR}(\theta)$ \Comment{Update model}
\EndFor
\State \textbf{end for}
\end{algorithmic}
\end{algorithm}

\textbf{Positive Tile Refinement.}
For positive tiles which contain objects, our refinement involves a transformer-based process for deep feature extraction and image reconstruction. To be specific, for a given tile $T_{LR}\in\mathbb{R}^{8p\times 8p\times C_{LR}}$, the convolution layer first extract the shallow feature $F_S\in\mathbb{R}^{8p\times 8p\times C_f}$, where $C_f$ is the number of channels for features. Subsequently, a series of $K$ residual transformer blocks (RTBs), based on the Swin transformer architecture, are employed to derive deep features $F_D\in\mathbb{R}^{8p\times 8p\times C_f}$ as the following equations:
\begin{equation}
    F_i = RTB_i(F_{i-1}), i=1, 2, ..., K, 
\end{equation}
\begin{equation}
    F_D = Conv(F_{K})
\end{equation}
where $F_0$ is the input shallow feature $F_S$, $RTB_i$ denotes the $i$-th RTB, and $Conv$ is the final convolution layer.

\begin{figure}[t]
\centering
\includegraphics[width=0.9\columnwidth]{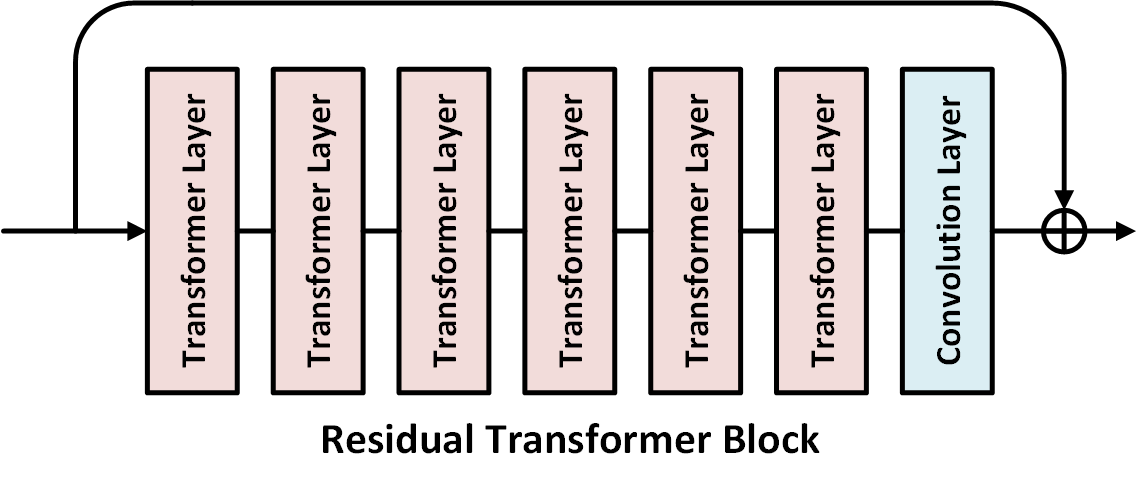} 
\caption{\textbf{Structure of RTB.} The incorporation of a skip connection and convolution layer in RTB contributes to improved feature learning capabilities.}
\label{fig:rtb}
\end{figure}

The keypoint is the design of RTB. Figure \ref{fig:rtb} presents the specific structure of RTB. Basically, it consists of a series of transformer layers and one convolution layer. Primarily, the inclusion of an additional convolutional layer at the end serves to optimize the transformer more effectively, yielding enhanced representations. This is due to the fact that direct similarity comparisons across all tokens often introduce redundancy, as evidenced in various works\cite{zhang2018image, liang2021swinir, li2023uniformer, wu2021cvt, xiao2021early}.
Secondly, the skip connection within the RTB establishes a link between features at different levels and the image reconstruction block. This facilitates the aggregation of features from diverse levels, promoting the integration of multi-level information. Additionally, we adopt distinct attention blocks proposed in \cite{chen2023activating} to activate more pixels for high-resolution image reconstruction.

After obtaining both the shallow feature $F_S$ and deep feature $F_D$, we merge them to reconstruct HR tiles using the following equation,
\begin{equation}
    T_{HR} = IR(F_S+F_D)
\end{equation}
where $IR$ is the image reconstruction block. By transmitting the shallow feature containing low-frequency information and the deep feature which highlights high-frequency details via a long skip connection,  this module effectively concentrates on reconstructing high-frequency information. And in this block, the sub-pixel convolution layer \cite{shi2016real} is employed to upsample the feature.

\textbf{Negative Tile Refinement.}
Since the intricate refinement for the background introduce some irrelevant or even detrimental features which degrade the image quality for downstream tasks, we remove the transformer-based feature extraction block and directly reconstruct the negative tiles with shallow features to obtain HR tiles with the same resolution as refined positive tiles.

After obtaining HR tiles via both PTR and NTR, we consolidate them to produce the refined HR image.

\subsection{Loss function}
For the TS module, the loss is computed using cross-entropy. With the introduction of a pyramid structure aimed at reducing false negative predictions (FN) and ensuring the selection of all positive patches, the losses from the three scales are combined to formulate the final loss as follows:

\begin{equation}
    \mathcal{L}_{TS} = \sum_{i=1}^3(-\boldsymbol{y}_ilog(\boldsymbol{s}_i) - (1-\boldsymbol{y}_i)log(1-\boldsymbol{s}_i))
\end{equation}

For the TR module, we leverage $L_1$ loss to quantify the discrepancy between the generated high-resolution image $I_{HR}$ and the ground truth high-resolution image $I_{GT}$ as follows:
\begin{equation}
    \mathcal{L}_{TR} = ||I_{HR}-I_{GT}||_1
\end{equation}
Finally, the loss function of SSR is the weighted sum of these two modules and the formulation is expressed as,
\begin{equation}
    \mathcal{L}_{SSR} = \mathcal{L}_{TS} + \alpha \mathcal{L}_{TR}
\end{equation}
where $\alpha$ is a hyper-parameter to adjust the weight of TS module in training.

\section{Experimental Results}

\subsection{Datasets and Experimental Settings}

We evaluate the effectiveness of our SSR model on three datasets, BDD100K \cite{bdd100k}, COCO 2017 \cite{lin2014microsoft} and MSRA10K \cite{ChengPAMI}. These datasets provide segmentation or object detection labels, enabling us to generate tile classification labels for tile selection. 
BDD100K is the most complex dataset focused on autonomous driving, containing approximately 100,000 high-quality images with an original resolution of $1280\times720$. Each image encompasses various objects, including vehicles, pedestrians, and eight other object categories.
COCO2017 comprises around 100,000 training images and 5,000 validation images. Each image encompasses multiple objects spanning 80 different classes. MSRA10K provides 10000 images and their segmentation labels. 

For all datasets, we resize them into low-resolution (LR) images with resolution $256\times256$, and high-resolution (HR) images with resolution $1024\times1024$ by bicubic interpolation. The embedding dimension of the Tile Selection (TS) module is 96 while it's 180 for Tile Refinement (TR) module. We set the learning rates to 0.00001. Each TL utilizes a depth of 2, a window size of 7, and 3 attention heads. We employ a patch size of 2 for embedding, which corresponds to tile sizes of $16\times16$, $32\times32$, and $64\times64$, yielding tile labels of $4\times4$, $8\times8$, and $16\times16$ respectively. The weight parameter $\alpha$ for the loss function is set to 1. The number of RTB is 6. All experiments are conducted for 50 epochs on two Linux servers, each equipped with two NVIDIA A6000 GPUs.

\subsubsection{Evaluation Metrics.}
Besides the common metrics for image quality, like Peak Signal-to-Noise Ratio (PSNR), Structural Similarity Index Measure (SSIM), Fréchet Inception Distance (FID), and Kernel Inception Distance (KID), we introduce an additional metric named Kernel YOLO Distance (KYD) to demonstrate the robustness of our approach.
To evaluate FID, we employ a pre-trained image classification model, Inceptionv3, to generate features and compare the distributions of two image sets. For KYD, we focus on the comparison of features for object detection (OD) by utilizing a pre-trained YOLOv8 model to generate features and compute the kernel distance similarly to KID.

Furthermore, we assess the performance of the TS module using True Positive Rate (TPR) and the maximum F1 score (maxF) for accuracy, along with the average number of selected tiles for computation efficiency.

\subsection{Tile Selection Results}
For Tile Selection (TS), we explore various  configurations to identify the optimal setup for our objectives. We adjusted the patch size of the first embedding layer, which consequently influenced the number of transformer layers (TL) in the network. Specifically, by setting the patch size to 2, 4, or 8, we achieved 5, 4, or 3 TLs respectively. We also experimented with different loss functions, including pyramid labels at various scales and a single label. Additionally, we introduced a max block at the end of the TS module to consolidate three predictions by selecting the maximum. The results, presented in Table \ref{tab:selection}, highlighted several key insights. Firstly, increasing the number of transformer layers enhanced performance, albeit at the cost of increased computation. Secondly, the pyramid structure consistently yielded better outcomes. Lastly, while the max block improved selection results, it also introduced additional computational overhead for the refinement module.

\begin{table}[!t]
    \centering
    \begin{threeparttable}
    \begin{tabularx}{\columnwidth}{p{2.3cm} |  c  c | c c }
    \toprule
    \textbf{Model} & \textbf{TPR} & \textbf{maxF}  & \textbf{\#Tiles} & \textbf{\#MACs}\\
    \midrule
    B (w/o pyramid) & \textbf{0.9137} &\textbf{0.8807} & \textbf{62\%} & 563.21M\\
    S (w/o pyramid) & 0.9086 &0.8522 & 65\% & 155.30M\\
    T (w/o pyramid) & 0.8973 & 0.8225 & 66\% & \textbf{53.11M}\\
    \midrule
    B (w/ pyramid) & \textbf{0.9341} & \textbf{0.8967} & \textbf{62\%} &563.21M\\
    S (w/ pyramid) & 0.9048 & 0.8327 & 66\% &155.30M\\
    T (w/ pyramid) & 0.8855 & 0.8384 & \textbf{62\%}  &\textbf{53.11M}\\
    \midrule
    B (w/ max) &\textbf{0.9692} & \textbf{0.8893} & \textbf{77\%} & 563.21M \\
    S (w/ max) & 0.9681 & 0.8572 & 79\% & 155.30M\\
    T (w/ max) & 0.9666 & 0.8400 & 81\% & \textbf{53.11M}\\
    \bottomrule
    \end{tabularx}
    \caption{\textbf{Results of TS module.} Encode images with 5, 4, or 3 transformer layers, corresponding to TS-Base (B), TS-Small (S) and TS-Tiny (T) in this table. More layers provide better performance with increasing computation. The adoption of the pyramid structure for tile selection across different scales improves the network's effectiveness. The integration of the max block further enhances the selection.}
    \label{tab:selection}
    \end{threeparttable}
\end{table}

\begin{table*}[!htb]
    \centering       
    \begin{threeparttable}
    \begin{tabularx}{\textwidth}{p{2.20cm} | c c c c c | c c c | c c c }
        \toprule
         \multirow{2}{*}{\textbf{Method}} & \multicolumn{5}{c|}{\textbf{BDD100K}}  & \multicolumn{3}{c|}{\textbf{COCO 2017}} & \multicolumn{3}{c}{\textbf{MSRA10K}} \\
         \cline{2-12}
          & \textbf{PSNR}$\uparrow$ & \textbf{SSIM}$\uparrow$  & \textbf{FID}$\downarrow$ & \textbf{KID}$\downarrow$ & \textbf{KYD}$\downarrow$ & \textbf{SSIM}$\uparrow$ & \textbf{FID}$\downarrow$ & \textbf{KYD}$\downarrow$ & \textbf{SSIM}$\uparrow$ & \textbf{FID}$\downarrow$ & \textbf{KYD}$\downarrow$\\
         \midrule
         SwinIR & 30.08 & 0.8757 & 24.87 & 0.0104 & 0.0092 & 0.8420 & 8.483 & 0.0036 & 0.9441 & 9.502 & 0.0035 \\
         HAT & \textbf{30.09} & 0.8750 & 26.78 & 0.0115 & 0.0096 & 0.8391 & 10.03 & 0.0037 & 0.9473 & 9.577 & 0.0030  \\
         \textbf{SSR} (ours) & 29.91 & \textbf{0.8833}&\textbf{10.41}  & \textbf{0.0017}& \textbf{0.0041}& \textbf{0.8512} & \textbf{2.844}&\textbf{0.0006} & \textbf{0.9653} & \textbf{1.905} & \textbf{0.0002} \\
         \bottomrule
    \end{tabularx}
\end{threeparttable}
\caption{\textbf{Comparison of SR results.} The table presents the outcomes for a 4$\times$ upscaling factor. For SwinIR and HAT, we utilize the official tile-based implementation of HAT for equitable assessment. Both SwinIR and HAT demonstrate results inferior to our SSR in terms of image quality across all three datasets. Generated images by SSR can perform better for downstream tasks with lower FID and KYD.}
    \label{tab:x4 results}
\end{table*}

\subsection{Super-Resolution (SR) results}
\begin{figure}[hbt]
\centering
\includegraphics[width=0.99\columnwidth]{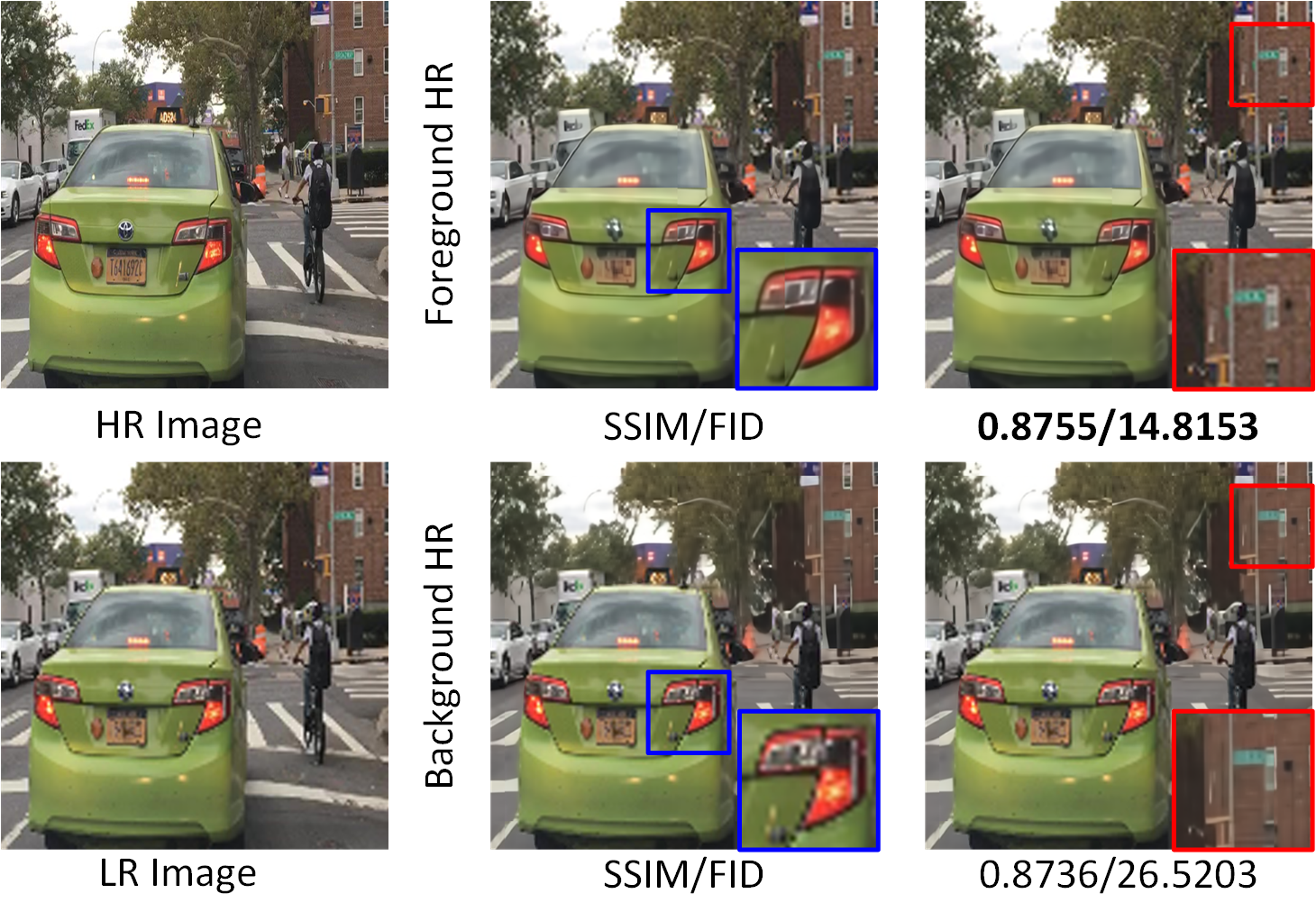} 
\caption{By replacing either the background or foreground pixels with bicubic-interpolated upscaled images, we effectively simulate the removal of high-frequency details from these regions. The results illustrate that the high-frequency information of background introduced by SR module impairs the generated image quality.}
\label{fig:replace}
\end{figure}

As shown in figure \ref{fig:intro}, by comparing the feature visualization results of reconstructed images by SR methods with ground truth HR images, we found the high-frequency refinement of conventional methods for background adversely impacts the image quality.  To substantiate this observation,  we devised two new datasets: one by replacing foreground pixels and the other by substituting background pixels with upscaled images via bicubic interpolation. This simulation aimed to explore the impact of high-frequency removal. Upon evaluating the image quality of these datasets, we discerned an intriguing finding. Even the mere replacement of pixels lacking high-frequency information for the background led to enhanced image quality. Figure \ref{fig:replace} graphically illustrates this comparison. The top example juxtaposes generated foreground via SR with coarse background upscaled by bicubic interpolation, while the bottom image maintains high-frequency details exclusively in the background. The top dataset exhibited superior image quality, as indicated by lower FID and higher SSIM. This quantitative assessment provides further support for our SSR design.

\begin{figure*}[t]
\centering
\includegraphics[width=\textwidth]{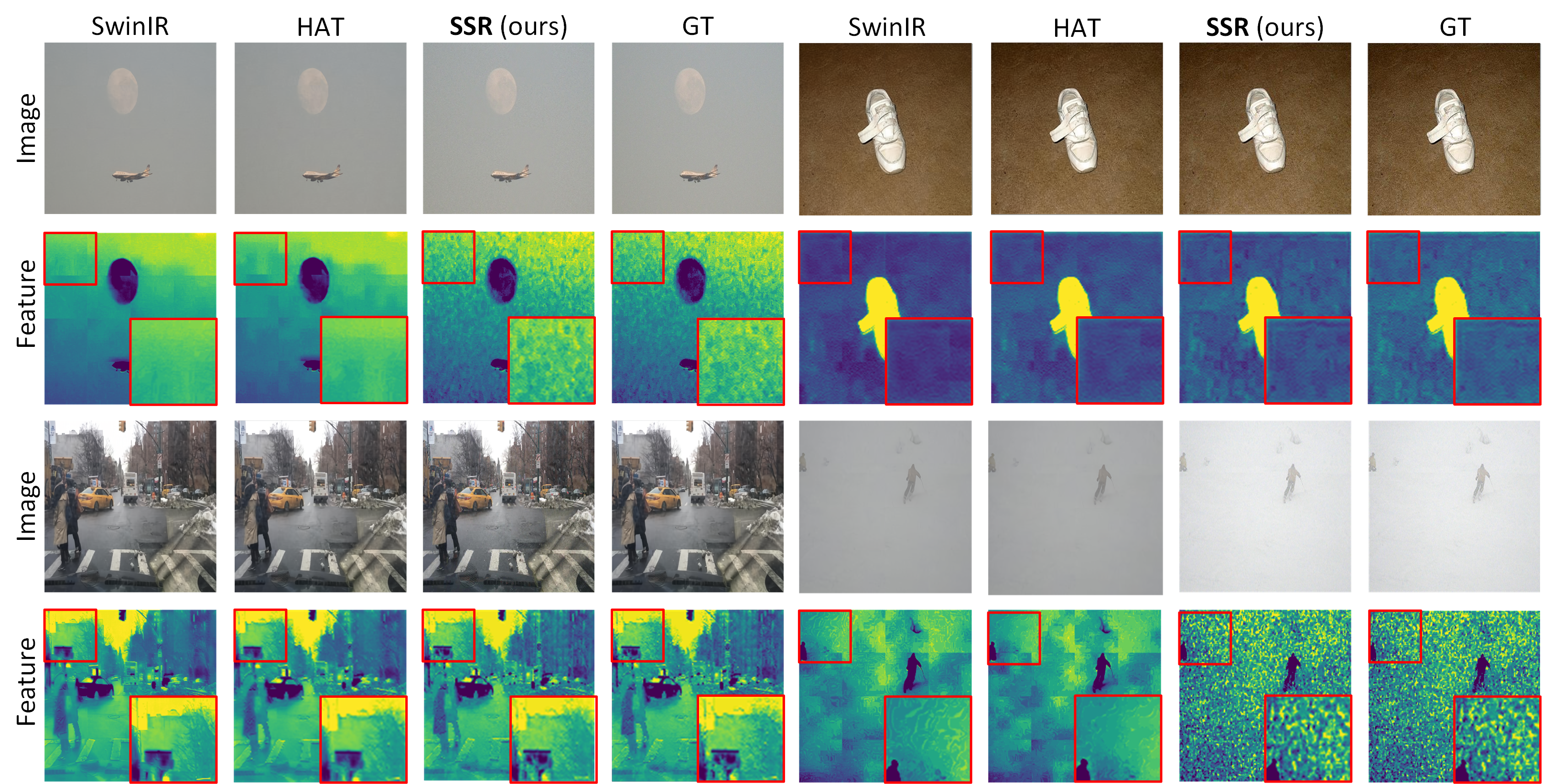} 
\caption{\textbf{Visual comparison with the state-of-the-art methods for $4\times$ upscaling.} We generate features with Inceptionv3 model and visualize them. The features extracted from SSR images exhibit a higher degree of similarity to the GT features.}
\label{fig:x4}
\end{figure*}

\subsubsection{Quantitative results.}
Table \ref{tab:x4 results} shows the quantitative comparison of our approach with the state-of-the-art methods \cite{liang2021swinir, chen2023activating} on three different datasets, considering a 4$\times$ upscale of the input images. The results clearly demonstrate the consistent superiority of our proposed method, SSR, in terms of various evaluation metrics. This robust performance underlines SSR's potential to generate images with improved features for downstream tasks such as image classification and object detection, as indicated by its lower FID and KYD.

In Table \ref{tab:x2x3}, we extend the comparison to 2$\times$ and 3$\times$ magnification factors on BDD100K.  Given SwinIR's challenges on this dataset, we focus on contrasting HAT with our method. The results showcased in this table demonstrate SSR's proficiency across different scales.

\begin{table}[!bt]
    \centering
    \begin{threeparttable}
    \begin{tabularx}{\columnwidth}{p{1.56cm} | c | c  c c c }
    \toprule
    \textbf{Method} & \textbf{Scale} & \textbf{SSIM} & \textbf{FID} & \textbf{FID} & \textbf{KYD}\\
    \midrule
     HAT & $\times2$ & 0.8764 & 27.58 &0.0123 & 0.0092\\
    \textbf{SSR }(ours) & $\times2$ &\textbf{0.8890} & \textbf{10.28} & \textbf{0.0016} & \textbf{0.0037}\\
    \midrule
    HAT &$\times3$ & 0.8768 & 30.88 & 0.0150 & 0.0096\\
    \textbf{SSR }(ours) &$\times3$ & \textbf{0.8918} & \textbf{10.80} & \textbf{0.0017}& \textbf{0.0027}\\
    \bottomrule
    \end{tabularx}
    \caption{\textbf{Quantitative comparison with HAT for 2$\times$ and 3$\times$ up-sampling.} SSR consistently outperforms HAT across all evaluation metrics, demonstrating its superiority in refining images at various scales.}
    \label{tab:x2x3}
    \end{threeparttable}
\end{table}

\subsubsection{Visual comparison.}
Figure \ref{fig:x4} illustrates the visual comparison across three datasets. Besides the ground truth (GT) images and the images generated by HAT and our SSR approach, we also present the extracted features from these images. The features derived from our SSR-generated images exhibit a closer resemblance to GT images, emphasizing the fidelity of our approach.

\subsubsection{Efficiency.}
Actually, with more tiles selected by the max block, the generated image quality remains quite similar, as shown in Table \ref{tab:max}. So, we opt not to include this block in our final algorithm.
For our SSR, only positive tiles are directed through the expensive SR module. In Table \ref{tab:computation}, we summarize the total number of parameters and computation cost for three different datasets. It underscores the efficiency of SSR with about 40\% computation reduction.

\begin{table}[!bt]
    \centering
    \begin{threeparttable}
    \begin{tabularx}{\columnwidth}{p{1.3cm} | c | c  c  cc }
    \toprule
    \textbf{Method} & \textbf{\#Tiles} & \textbf{SSIM} & \textbf{FID} & \textbf{KID}& \textbf{KYD}\\
    \midrule
     w/ max & 62\% & 0.8833 & 10.41 & 0.0017 & 0.0041\\
    w/o max & 77\%& 0.8835 & 10.62 & 0.0017 & 0.0041\\

    \bottomrule
    \end{tabularx}
    \caption{With the max block, the visual quality of images generated by SSR can be slightly improved with additional 15\% computation overhead.}
    \label{tab:max}
    \end{threeparttable}
\end{table}

\begin{table}[!bt]
    \centering
    \begin{threeparttable}
    \begin{tabularx}{\columnwidth}{p{1.6cm} | c | c  c c }
    \toprule
    \textbf{Dataset} & \textbf{Method} & \textbf{\#Tiles} & \textbf{\makecell{\#Params\\(M)}} & \textbf{\makecell{\#MACs\\(G)}}\\
    \midrule
     \multirow{2}{*}{BDD100K} & HAT & - & 1675.99 & 191.06\\
      & \textbf{SSR} & 62\% & \textbf{1233.13}& \textbf{119.40}\\
      \midrule
    \multirow{2}{*}{COCO 2017} & HAT & - & 1675.99 & 191.06\\
      & \textbf{SSR} & 65\% & \textbf{1233.13}& \textbf{125.15}\\
      \midrule
    \multirow{2}{*}{MSRA10K} & HAT & - & 1675.99 & 191.06\\
      & \textbf{SSR} & 57\% & \textbf{1233.13}& \textbf{109.82}\\
    \bottomrule
    \end{tabularx}
    \caption{\textbf{Computation cost comparison.} Compared to HAT, SSR achieves about 40\% computation reduction for all three datasets. All metrics are evaluated for a $256\times256$ input.}
    \label{tab:computation}
    \end{threeparttable}
\end{table}

\subsection{Ablation Study}
\subsubsection{Pre-training with ImageNet.}
\begin{figure}[!bt]
\centering
\includegraphics[width=0.99\columnwidth]{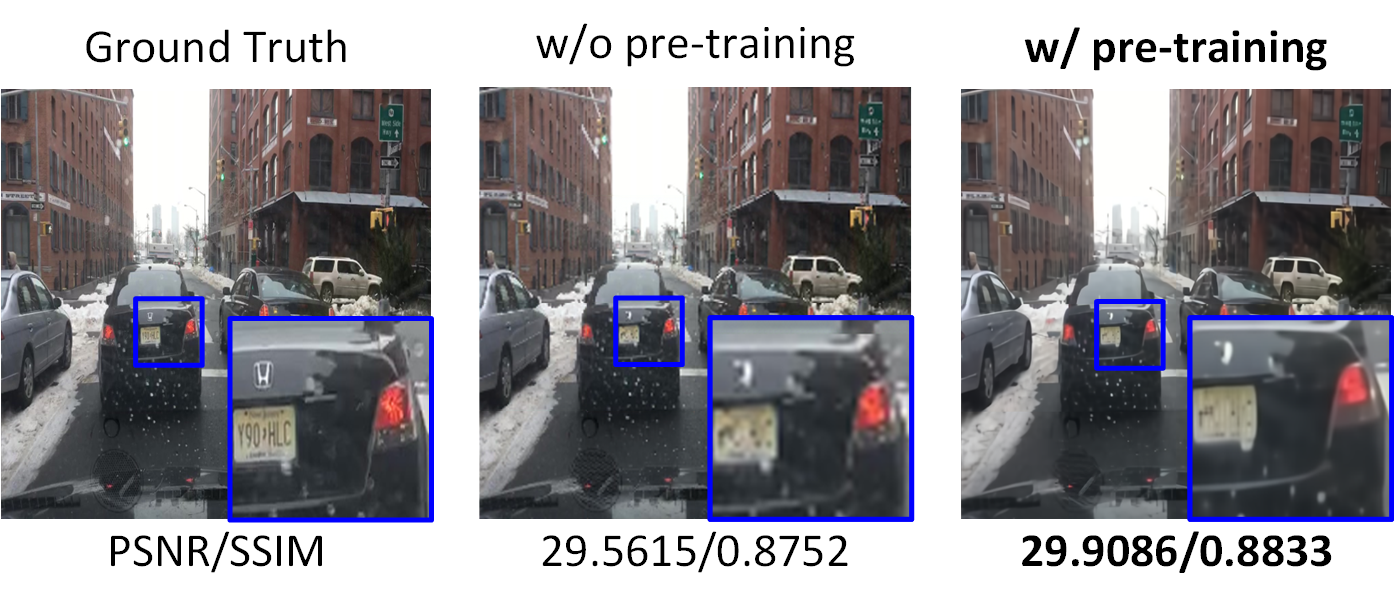} 
\caption{With pre-training, SSR can generate better images compared to the outcomes without pre-training.}
\label{fig:pretrain}
\end{figure}

Similar to other computer vision work, SSR can substantially benefit from a pre-training strategy employing a large dataset, such as ImageNet. A comparison of results obtained with and without pre-training highlights the enhancement in image quality, with PSNR scores improving from 29.56 to 29.91. Figure \ref{fig:pretrain} presents one example for visual comparison.

\subsubsection{Negative Tile Refinement.}
For background pixels, high-frequency information recovery is unnecessary. We experimented with two different designs: the first employs a single Upsample layer, while the second retains the convolution layers from the Positive Tile Refinement (PTR) path and omits the transformer blocks. The results are shown in Table \ref{tab:negative}. Adding convolution layers in NTR marginally enhances image quality. This enhancement can be attributed to the compensation for a few positive tiles that might be missed by the selection module. However, this improvement comes at the cost of increased 19.2\% computational load. Therefore, it is a trade-off between the computation and image generation performance.

\begin{table}[!bt]
    \centering
    \begin{threeparttable}
    \begin{tabularx}{\columnwidth}{p{1.27cm} | c  c  c  cc }
    \toprule
    \textbf{Method} & \textbf{PSNR} & \textbf{SSIM}& \textbf{FID} & \textbf{KID}& \textbf{KYD}\\
    \midrule
     w/o conv & 29.91 & 0.8833 & 10.41 & 0.0017 & \textbf{0.0041}\\
    \textbf{w/ conv} & \textbf{30.03} & \textbf{0.8863} & \textbf{9.56} & \textbf{0.0013} & 0.0042\\
    \bottomrule
    \end{tabularx}
    \caption{With convolution layers in the NTR path, SSR showcases an improvement in image generation.}
    \label{tab:negative}
    \end{threeparttable}
\end{table}

\section{Conclusion}
In this paper, we delve into the cause behind the image quality gap between generated images through conventional super-resolution (SR) techniques and high-resolution (HR) ground truth images. Our investigation reveals that the high-frequency refinement of background pixels undermines the overall image feature generation for downstream tasks. To address this challenge and concurrently reduce computational overhead, we introduce a novel algorithm termed Selective Super-Resolution (SSR). By leveraging a cost-efficient transformer-based network, we partition the input low-resolution (LR) image into non-overlapping tiles, assessing the presence of objects within each tile. By exclusively refining the high-frequency characteristics of object-containing tiles, we improve visual quality with a lower computation cost. Our approach's superior performance is substantiated across three distinct datasets employing diverse evaluation metrics. Specifically, experiments on BDD100K exhibit an improvement of FID from 26.78 to 10.41 with 40\% computation reduction.

\section*{Acknowledgments}
This work is supported by the Center for the Co-Design of Cognitive Systems (CoCoSys), one of seven centers in Joint University Microelectronics Program 2.0 (JUMP 2.0), a Semiconductor Research Corporation (SRC) program sponsored by the Defense Advanced Research Projects Agency (DARPA).

\bibliography{aaai24}

\end{document}